\documentclass[cameraready]{Interspeech}

\usepackage[sidenotes,mnwidth=1.8cm]{mnotes}
\Mnewauthor[SL]{SL}{red}
\Mnewauthor[SF]{SF}{orange}
\Mnewauthor[EL]{EL}{cyan}

\renewcommand{\mnSL}[1]{}
\renewcommand{\mnSF}[1]{}
\renewcommand{\mnEL}[1]{}

\title{Can We Trust LLMs for Mental Health Screening?\\
Consistency, ASR Robustness, and Evidence Faithfulness}

\author[affiliation={1}, correspondingauthor]{Erfan}{Loweimi}
\author[affiliation={1}]{Sofia}{de la Fuente Garcia}
\author[affiliation={2}]{Samira}{Loveymi}
\author[affiliation={3}]{Hadi}{Daneshvar}
\author[affiliation={1}]{Saturnino}{Luz}

\address{
    $^1$ Usher Institute, Edinburgh Medical School, University of Edinburgh, Edinburgh, UK \\
    $^2$ Department of Computer Engineering, Ahvaz Campus, Islamic Azad University, Ahvaz, Iran\\
    $^3$ School of Health and Social Care, Edinburgh Napier University, Edinburgh, UK
}

\email{e.loweimi@ed.ac.uk}

\keywords{Mental health, LLM, HADS, depression, anxiety, ASR, consistency, reliability, keywords}

\usepackage{multirow}

\newcommand{\scc}{\rho_s}

\graphicspath{{figs/}}

\begin{document}

\maketitle

\begin{abstract}
  LLMs can estimate Hospital Anxiety and Depression Scale (HADS)
  scores from speech in a zero-shot manner, but clinical deployment
  requires reliability across three dimensions: intra-model
  consistency, ASR robustness, and evidence faithfulness. We evaluate
  three LLMs (Phi-4, Gemma-2-9B, and Llama-3.1-8B) on 111 English-speaking participants
  using ground-truth transcripts and three Whisper ASR variants
  (Large, Medium, Small), with three independent runs per
  model–condition pair. We find that (i) Phi-4 and Gemma-2-9B achieve
  excellent intra-model consistency ($\text{ICC} > 0.89$) with minimal
  degradation under ASR; (ii) Llama-3.1-8B shows ASR-fragile
  consistency, with ICC dropping from 0.82 to 0.36 at 10\% WER; (iii)
  predictive validity is largely preserved under ASR for robust
  models; and (iv) keyword groundedness exceeds 93\% for Phi-4 and
  Gemma-2-9B but falls to 77--81\% for Llama-3.1-8B. Inter-model
  keyword agreement is far lower than score-level agreement, revealing
  a score–evidence dissociation with implications for clinical
  interpretability.
\end{abstract}

\section{Introduction}
\label{sec:intro}

Mental health disorders such as anxiety and depression impact millions
worldwide each year~\cite{salari2020}. Although early detection is
critical for effective intervention~\cite{mcgorry2018}, traditional
screening relies on clinical interviews~\cite{vonkorff1987} or the
Hospital Anxiety and Depression Scale (HADS)~\cite{zigmond1983},
approaches that are resource-intensive, subjective, and difficult to
administer at scale~\cite{newson2020}. While HADS scores are
obtainable via questionnaire, self-report relies on active
participation, introspective access, and willingness to disclose,
assumptions that fail for populations affected by stigma, cognitive
decline, or language barriers.\mnSL{This poses a bit of a dilemma for
  model assessment against a HADS ``gold-standard''. Maybe this should
be pointed out as a limitation of the approach adopted here. However,
perhaps point this out and discussing it should be left for a
longer, more clinically-orientated journal paper.}

Speech-based estimation offers a scalable alternative by sidestepping
these barriers, enabling passive screening embedded in routine
interactions and longitudinal monitoring free of questionnaire
fatigue. More fundamentally, as the field moves toward integrating
language models into clinical workflows, reliable \textit{screening}
is a practical prerequisite for downstream applications such as
treatment monitoring.
Accurate HADS estimation therefore serves as a validation
gate for establishing broader clinical utility.

A rich body of work has established that language carries reliable
markers of psychological states: depression manifests through negative
emotion words, first-person pronouns, and reduced lexical
diversity~\cite{rude2004,tausczik2010}, while anxiety is associated
with hedging, uncertainty expressions, and hesitation
markers~\cite{stade2023}.  These insights have driven text-based
screening from classical machine learning on social
media~\cite{guntuku2017,eichstaedt2018} and clinical interviews
through deep learning approaches~\cite{malhotra2022,calvo2017}.  In
the speech domain, acoustic features have been successfully used for
depression and emotion recognition~\cite{cummins2015,wu2011,low2020},
and multimodal systems combining audio and text further improve
detection~\cite{alhanai2018}.

Large language models (LLMs), whose zero-shot capabilities were
established by Brown et al.~\cite{brown2020}, have extended these
approaches to estimating psychological states without task-specific
training.  LLM-based systems have shown
promise for depression detection from social
media~\cite{lan2024,eichstaedt2018}, clinical
dialogues~\cite{tao2023,xu2025}, and personal
narratives~\cite{lho2025}, with systematic reviews confirming their
growing role~\cite{guo2024,liu2025}.  Zero-shot mental health
classification can approach supervised
baselines~\cite{lamichhane2023}, and interpretability-focused
prompting strategies further improve clinical utility~\cite{yang2024}.
It was also demonstrated that diverse LLMs can estimate HADS scores
from speech transcripts with moderate validity (Spearman $\scc = 0.38$--$0.58$)~\cite{loweimi2025,loweimi2025jstsp},
while Roy et al.~\cite{roy2025} showed zero-shot diagnostic
performance comparable to fine-tuned models.  However, these studies
evaluate only predictive accuracy, and methodological concerns
persist~\cite{ernala2019}: \textit{reliability}, \textit{robustness},
and \textit{interpretability} remain largely unexamined, yet are
indispensable for clinical adoption.

Three specific gaps motivate this work.  First, LLM inference is
inherently stochastic, yet the reproducibility of clinical predictions
across repeated runs has received little attention.  Second, while
LLMs can cite evidence for their predictions, chain-of-thought
explanations may be unfaithful to the model's actual decision
process~\cite{turpin2024}, and hallucinated clinical evidence could
mislead clinicians~\cite{ji2023}, making evidence groundedness
verification imperative~\cite{joyce2023}.  Third, practical deployment
requires automatic speech recognition (ASR), yet modern ASR systems
like Whisper~\cite{radford2023} systematically delete disfluencies and
hesitation markers in clinical
speech~\cite{delafuente2021}, precisely the features LLMs may rely on
for psychological assessment, and the interaction between ASR errors
and downstream clinical predictions remains unexplored.

We address these gaps with four research questions, whose systematic
investigation forms this paper’s core contributions:

\begin{enumerate}
\item Are LLM predictions stable across repeated runs, and does ASR
  affect this stability?
\item Does ASR degrade predictive validity (correlation with HADS
  ground truth)?
\item Are LLM-cited keywords grounded in the transcript, and is
  keyword evidence stable across runs and models?
\item Are there model-specific vulnerability patterns that inform
  deployment decisions?
\end{enumerate}

Having introduced the dataset, HADS instrument, and prompt design in
Section~\ref{sec:data}, we present the statistical analysis framework
underpinning our evaluation in
Section~\ref{sec:stats}. Section~\ref{sec:setup} describes the
experimental setup. Section~\ref{sec:results} presents results as well
as
discussion 
and Section~\ref{sec:conclusions} concludes the
paper.

\section{Data, HADS, and Prompt Design}
\label{sec:data}

\subsection{The PsyVoiD corpus}

We use the PsyVoiD corpus~\cite{psyvoid2023}, collected in Scotland (UK)
during the COVID-19 lockdown to investigate the relationship between spontaneous
English speech and psychological traits. The dataset comprises 111
participants (70 female, 41 male), aged 21--86 (median 62), 34 of whom (31\%)
report a prior clinical history of depression (not necessarily current).
Participants recorded one-to-two-minute spontaneous speech samples via personal devices in home environments,
describing their daily lives during lockdown, yielding transcripts with considerable length
variability (mean 152 words, SD = 77). Each participant completed the Hospital
Anxiety and Depression Scale (HADS) questionnaire~\cite{zigmond1983}
on the same day as the recording, providing paired ground-truth
anxiety and depression scores.
The study was approved by the NHS Health Research Authority (REC reference: 20/EM/0146); all participants provided informed consent.

\subsection{The HADS instrument}

HADS~\cite{zigmond1983} is a widely validated self-report
instrument~\cite{bjelland2002,herrmann1997}, yielding two subscale
scores: Anxiety (HADS-A) and Depression (HADS-D), each ranging from 0
to~21, with established clinical thresholds (0--7 normal, 8--10
borderline, 11--21 clinical). In PsyVoiD, HADS-A has a mean of 6.7 (SD = 4.9) and HADS-D has a mean of 5.0 (SD = 4.5).
While all three severity tiers are represented, the distribution is skewed toward the normal range, with relatively few participants in the clinical tier (HADS $\geq$ 11).
\mnSL{I've added the SD values based on the data I have (for some
  reason 110 participants only). Please double check.}

\subsection{Prompt design}
\label{sec:prompt}

\begin{figure}[t]
  \centering
  \includegraphics[width=\linewidth]{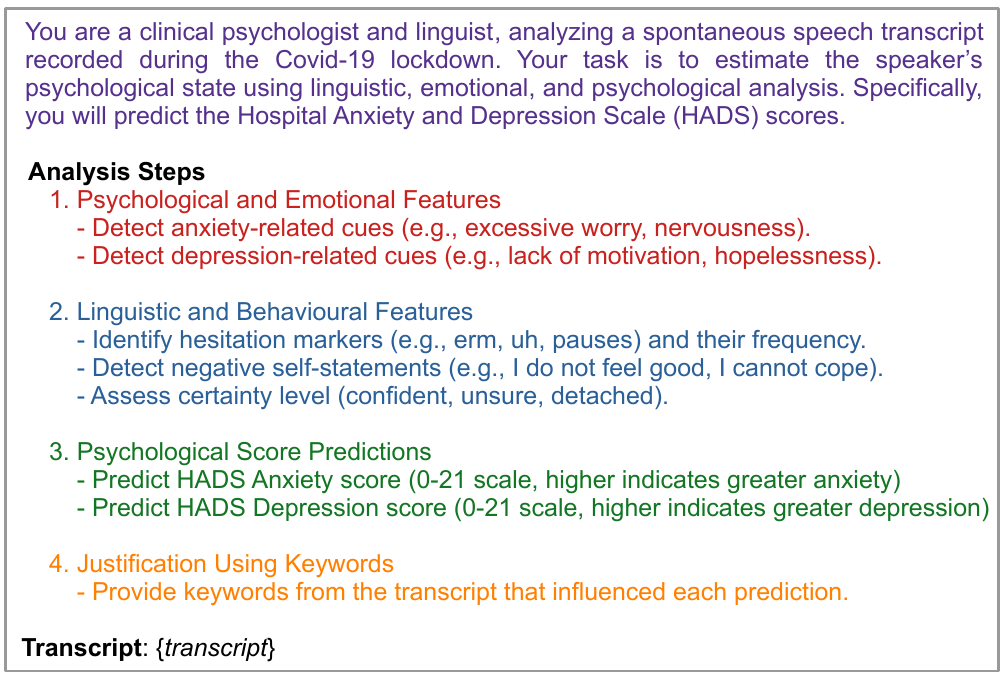}
  \vspace{-2mm}
  \caption{Zero-shot prompt used for HADS score estimation, integrating role specification, step-by-step decomposition, score prediction, and keyword justification. Adapted from~\cite{loweimi2025}.}
  \label{fig:prompt}
\end{figure}

To maximise reliability and interpretability, we designed a structured prompt informed by clinical psychology and linguistics expertise~\cite{loweimi2025}.
Figure~\ref{fig:prompt} shows the complete prompt.
It embodies four design principles:
\begin{itemize}
    \item \textit{role specification}: assigning the LLM a dual expert persona (clinical psychologist and linguist) to activate domain-relevant knowledge~\cite{wei2022,guo2024};

    \item \textit{step-by-step decomposition}: structuring the analysis into four sequential steps (psychological feature extraction, linguistic analysis, score prediction, keyword justification) to enforce chain-of-thought reasoning~\cite{brown2020,wei2022} before scoring;

    \item \textit{keyword justification}: requiring the model to cite specific transcript keywords, producing an auditable evidence trail and constraining predictions to be grounded in the input~\cite{joyce2023};

    \item \textit{structured JSON output}: requested via the prompt (not enforced through constrained decoding) to minimise free-text variability and facilitate automated parsing.
\end{itemize}

\section{Statistical Analysis Framework}
\label{sec:stats}

In this section we describe the measures used to evaluate the three reliability dimensions. All statistical tests are non-parametric, reflecting the ordinal nature of HADS scores.

\subsection{Intra-model consistency}
To assess whether repeated LLM inference yields stable predictions, we use the Friedman test to detect systematic inter-run differences, and ICC(3,1) (two-way mixed, single measures, consistency~\cite{shrout1979}) to quantify inter-run agreement. ICC thresholds follow Koo and Li~\cite{koo2016}: $<$0.50 poor, 0.50--0.75 moderate, 0.75--0.90 good, $>$0.90 excellent.

\subsection{Predictive validity and ASR robustness}
To evaluate predictive validity and quantify the impact of ASR errors on LLM-based HADS estimation, we use the following complementary measures:

\begin{itemize}
\item \textit{Spearman’s rank correlation} ($\scc$) measures the monotonic association between LLM predictions, computed as the mean of three inference runs, and HADS ground truth. It captures rank-order validity without assuming linearity or normality.
\item \textit{Wilcoxon signed-rank test} assesses systematic distributional differences between predicted and ground-truth HADS scores. A significant result ($p < 0.05$) indicates consistent over- or under-estimation by the model.
\item \textit{ASR robustness metrics} directly quantify the effect of transcription errors by comparing ground-truth-based and ASR-based predictions for each subject using mean absolute error (MAE) and paired $\scc$.
\end{itemize}

\subsection{Keyword evidence analysis}
To evaluate the faithfulness and stability of the transcript evidence cited by LLMs in support of their predictions, we analyse keyword behaviour along three complementary dimensions:

\begin{itemize}
\item \textit{Groundedness} measures whether cited keywords are present in the source transcript and is assessed using exact case-insensitive substring matching, supplemented by fuzzy matching based on the Levenshtein partial ratio ($\geq$ 80) implemented via \texttt{rapidfuzz}.

\item \textit{Intra-model keyword stability} is quantified using Jaccard similarity between keyword sets across three independent runs for each subject, capturing whether a model consistently cites the same evidence when re-run under identical conditions.

\item \textit{Inter-model keyword agreement} is measured using Jaccard similarity across different models for each subject, quantifying the extent to which distinct LLMs ground their predictions in the same transcript evidence.
\end{itemize}

\section{Experimental Setup}
\label{sec:setup}

\subsection{LLM configurations}

We evaluate three open-weight instruction-tuned LLMs, selected to span distinct model families, training pipelines, and parameter scales:
\text{Phi-4} (14.7B, Microsoft~\cite{abdin2024}),
\text{Gemma-2-9B} (9B, Google~\cite{gemma2024}), and
\text{Llama-3.1-8B-Instruct} (8B, Meta~\cite{grattafiori2024}).
Following~\cite{loweimi2025,loweimi2025jstsp}, each model receives the zero-shot prompt (Section~\ref{sec:prompt}) containing the transcript and produces HADS-A and HADS-D scores, and supporting keywords.
For each model--transcript combination, we perform three independent inference runs with default sampling (temperature~$\approx$~0.7) to quantify intra-model variability.
We deliberately retain stochastic sampling rather than setting temperature to~0, because characterising intra-model variability under realistic deployment conditions is itself a central research question of this study.

\subsection{Transcript conditions}

Each model is evaluated under four transcript conditions:
(1)~GT: ground-truth human transcription (WER~=~0\%);
(2)~W-Large: Whisper Large-v3 (WER~=~8.6\%);
(3)~W-Medium: Whisper Medium-en (WER~=~9.3\%);
(4)~W-Small: Whisper Small-en (WER~=~10.1\%).
The three Whisper variants were chosen to span a range of model capacities and WER levels within a single ASR family, isolating the effect of transcription quality from architectural differences.
Note that these WER values reflect Whisper's performance on spontaneous elderly speech, which is higher than the WER typically reported on read-speech benchmarks.
The Whisper error profile on PsyVoiD is deletion-heavy (4.1--4.6\% deletion rate)~\cite{loweimi2025}, preferentially removing hesitation markers and disfluencies common in elderly spontaneous speech~\cite{delafuente2021}.
Each Whisper variant was run once to produce a single deterministic transcript per speaker; the three inference runs vary only the LLM sampling, not the ASR output.
This yields 3 models $\times$ 4 conditions $\times$ 3 runs = 36 experimental configurations.

\section{Experimental Results and Discussion}
\label{sec:results}

\subsection{Consistency across runs}
\label{sec:intra}

Table~\ref{tab:consistency} reports ICC(3,1) and Friedman $p$-values
across all model--condition--subscale combinations.  Phi-4 and
Gemma-2-9B show excellent consistency with minimal degradation across
ASR conditions.  Phi-4's ICC ranges from 0.890 to 0.925 across all
conditions and subscales ($\Delta_{\max}$\,=\,0.035).  Gemma-2-9B
achieves even higher ICC (0.947--0.964), with a maximum
cross-condition variation of only 0.017.  Neither model shows
significant Friedman effects in any condition (all $p > 0.05$),
confirming that no run systematically deviates.%
\mnSL{I am not sure Friedman adds much to the analysis, but I guess it does not harm
  leaving it there.}



Llama-3.1-8B emerges as the critical outlier. Under GT transcripts,
its intra-model consistency is borderline acceptable (HADS-A: 0.753;
HADS-D: 0.818). Under ASR conditions, however, consistency degrades
sharply. At W-Small (10.1\% WER), ICC drops to 0.358 for HADS-A and
0.381 for HADS-D, falling below the ``poor'' reliability threshold of
0.50~\cite{koo2016}.%
\mnSL{It is surprising that this should happen, given that all runs
  under a given ASR regime receive the same noisy data. This peculiar
  interaction between ASR WER and consitency across runs might deserve
  closer scrutiny (though not necessarily in this paper).}
The Friedman test is also significant in this
condition ($p = 0.04$ for HADS-A; $p = 0.03$ for HADS-D), indicating
systematic differences across inference runs rather than random
sampling noise.
These results suggest that ASR errors interact with Llama-3.1-8B's
internal representations in a manner that amplifies stochastic
variability, whereas Phi-4 and Gemma-2-9B remain robust to such
perturbations. The tight coupling between score-level consistency and
keyword stability (Gemma-9B~$>$~Phi-4~$>$~Llama-8B; see
Section~\ref{sec:keywords}) further reinforces the conclusion that
reliable predictions cannot be achieved in the absence of stable and
consistent evidence generation.

\begin{table}[t]
\caption{Intra-model consistency: ICC(3,1) and Friedman $p$-value per model, transcript condition, and HADS subscale (A\,=\,Anxiety, D\,=\,Depression). Bold: ICC\,$\geq$\,0.85; $^\text{\textdagger}$: $p<0.05$.}
\label{tab:consistency}
\centering
\setlength{\tabcolsep}{3pt}
\scalebox{0.95}{
\begin{tabular}{@{}lccccc@{}}
\toprule
& & \multicolumn{4}{c}{ICC\,/\,Friedman $p$} \\
\cmidrule(l){3-6}
Model & HADS & GT & W-Large & W-Med. & W-Small \\
\midrule
\multirow{2}{*}{Phi-4}
  & A & \textbf{.898}\,/\,.43 & \textbf{.900}\,/\,.51 & \textbf{.890}\,/\,.78 & \textbf{.897}\,/\,.44 \\
  & D & \textbf{.925}\,/\,.17 & \textbf{.895}\,/\,.66 & \textbf{.914}\,/\,.85 & \textbf{.903}\,/\,.32 \\
\midrule
\multirow{2}{*}{Gemma-9B}
  & A & \textbf{.964}\,/\,.44 & \textbf{.955}\,/\,.58 & \textbf{.954}\,/\,.62 & \textbf{.955}\,/\,.12 \\
  & D & \textbf{.947}\,/\,.96 & \textbf{.948}\,/\,.46 & \textbf{.949}\,/\,.89 & \textbf{.953}\,/\,.52 \\
\midrule
\multirow{2}{*}{Llama-8B}
  & A & .753\,/\,.10 & .709\,/\,.65 & .728\,/\,.21 & .358\,/\,.04$^\text{\textdagger}$ \\
  & D & .818\,/\,.46 & .741\,/\,.70 & .697\,/\,.69 & .381\,/\,.03$^\text{\textdagger}$ \\
\bottomrule
\end{tabular}}
\end{table}

\subsection{Robustness to ASR quality}
\label{sec:asr}

\begin{figure}[t]
  \centering
  \includegraphics[width=\linewidth]{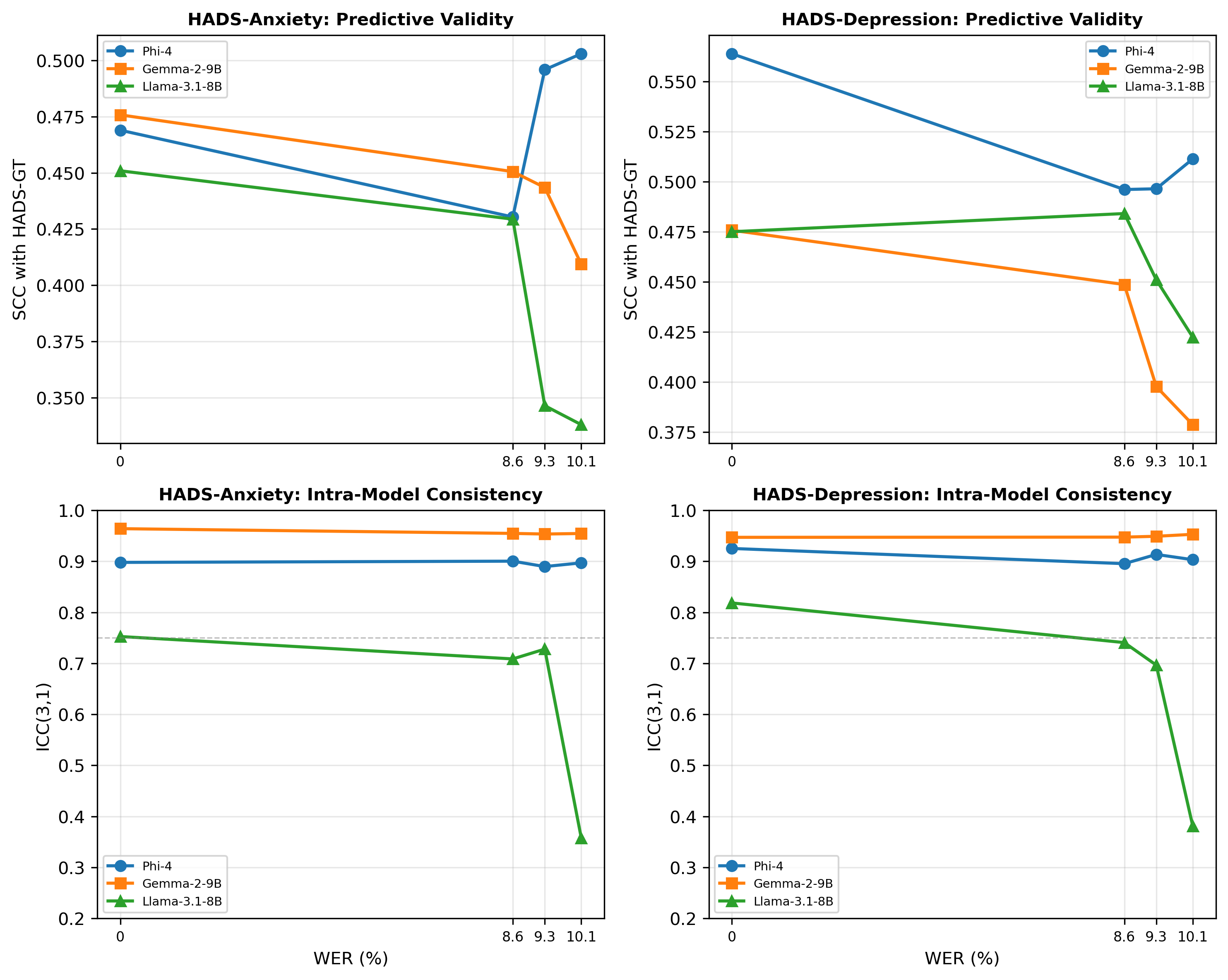}
  \caption{Top: Predictive validity (Spearman $\scc$ with HADS ground truth) vs WER. Bottom: Intra-model consistency (ICC) vs WER. Note the sharp ICC decline for Llama-3.1-8B at higher WER, contrasting the stability of Phi-4 and Gemma-2-9B.}
  \label{fig:combined}
\end{figure}

Table~\ref{tab:validity} and Figure~\ref{fig:combined} show Spearman
correlations between model predictions and HADS ground truth across
transcript conditions.  Phi-4 exhibits an unexpected pattern for
HADS-A: validity \textit{increases} from $\scc = 0.469$ (GT) to
$0.503$ (W-Small).  Whisper's deletion-heavy error profile removes
disfluencies (``erm,'' false starts) that Phi-4 over-interprets as
anxiety features, as confirmed by keyword frequency analysis where
``erm'' dominates Phi-4's HADS-A keyword list
(Figure~\ref{fig:keywords}).  Since PsyVoiD participants skew older (median age 62) and
disfluency patterns in elderly spontaneous speech correlate with arousal rather than
anxiety~\cite{delafuente2021}, their removal inadvertently corrects a
systematic bias, with ASR acting as a beneficial denoising filter for
this model.  For HADS-D, validity decreases mildly ($0.564 \to 0.511$,
$\Delta = 0.053$).  This paradox is model-specific: Gemma-2-9B shows a
gentle monotonic decline for both subscales (HADS-A:
$0.476 \to 0.409$; HADS-D: $0.476 \to 0.379$), and Llama-3.1-8B
experiences the steepest HADS-A decline ($0.451 \to 0.338$,
$\Delta = 0.113$), paralleling its ICC collapse.  Depression validity
is generally more resilient, consistent with depression markers being
more lexically robust than prosodically-mediated anxiety
cues~\cite{cummins2015,stade2023}.

Table~\ref{tab:gt_asr} provides the direct subject-level comparison
between GT-based and ASR-based predictions.  Phi-4 and Gemma-2-9B
produce GT-vs-ASR predictions with $\scc = 0.75$--$0.87$ and MAE of
0.49--1.00 points, suggesting
that for these models, ASR-induced shifts are small relative to the prediction scale.  Llama-3.1-8B shows substantially poorer agreement
(MAE up to 2.06 points, $\scc$ as low as 0.57), confirming that ASR
perturbations interact destructively with its inference.

Pairwise inter-LLM agreement ($\scc$ between models' mean predictions)
ranges from 0.62 to 0.87 (all $p < 10^{-4}$).  Phi-4--Gemma-9B
achieves the strongest agreement ($\scc = 0.87/0.84$ for HADS-A/D
under GT), while Llama-8B pairs are weakest under ASR
($\scc = 0.62$--$0.65$ at W-Small), mirroring its intra-model
instability.  These results highlight that model selection dominates
ASR condition as a source of variation: the reliability gap between
Phi-4/Gemma-9B and Llama-8B far exceeds the difference between any two
Whisper variants, and even smaller ASR models suffice ($\scc$
difference between W-Large and W-Small $\leq 0.07$ for reliable LLMs).

\begin{table}[t]
\caption{Predictive validity: Spearman $\scc$ between LLM predictions (mean of 3 runs) and HADS ground truth. All $p < 0.001$.}
\label{tab:validity}
\centering
\setlength{\tabcolsep}{3.5pt}
\begin{tabular}{@{}lccccc@{}}
\toprule
Model & HADS & GT & W-Large & W-Med. & W-Small \\
\midrule
\multirow{2}{*}{Phi-4}
  & A & 0.469 & 0.430 & 0.496 & \textbf{0.503} \\
  & D & \textbf{0.564} & 0.496 & 0.496 & 0.511 \\
\midrule
\multirow{2}{*}{Gemma-9B}
  & A & \textbf{0.476} & 0.451 & 0.443 & 0.409 \\
  & D & \textbf{0.476} & 0.449 & 0.398 & 0.379 \\
\midrule
\multirow{2}{*}{Llama-8B}
  & A & \textbf{0.451} & 0.429 & 0.346 & 0.338 \\
  & D & 0.475 & \textbf{0.484} & 0.451 & 0.422 \\
\bottomrule
\end{tabular}
\end{table}

\begin{table}[t]
\caption{GT-vs-ASR prediction agreement: MAE, Spearman $\scc$, and \% of predictions within 1~point. All use mean-of-3-runs.}
\label{tab:gt_asr}
\centering
\setlength{\tabcolsep}{2.5pt}
\begin{tabular}{@{}llccccc@{}}
\toprule
& & \multicolumn{2}{c}{MAE} & \multicolumn{2}{c}{Spearman $\scc$} & \\
\cmidrule(lr){3-4} \cmidrule(lr){5-6}
Model & ASR & A & D & A & D & \%$\leq$1 \\
\midrule
\multirow{3}{*}{Phi-4}
  & W-L & 0.87 & 0.87 & 0.810 & 0.873 & 73.1 \\
  & W-M & 0.74 & 0.77 & 0.753 & 0.820 & 78.8 \\
  & W-S & 0.81 & 0.75 & 0.780 & 0.843 & 80.7 \\
\midrule
\multirow{3}{*}{Gemma-9B}
  & W-L & 0.86 & 0.49 & 0.867 & 0.828 & 79.2 \\
  & W-M & 1.00 & 0.51 & 0.848 & 0.871 & 79.7 \\
  & W-S & 0.93 & 0.52 & 0.845 & 0.871 & 78.8 \\
\midrule
\multirow{3}{*}{Llama-8B}
  & W-L & 1.78 & 1.46 & 0.669 & 0.678 & 44.3 \\
  & W-M & 1.97 & 1.45 & 0.588 & 0.618 & 37.3 \\
  & W-S & 2.06 & 1.72 & 0.573 & 0.596 & 45.8 \\
\bottomrule
\end{tabular}
\end{table}

\subsection{Keyword groundedness}
\label{sec:keywords}

Beyond scores, each LLM outputs transcript keywords to justify predictions.
Table~\ref{tab:keywords} summarises three evidence dimensions.
Phi-4 and Gemma-2-9B achieve $>$93\% groundedness
, meaning fewer than 1 in 14 keywords are fabricated.
Llama-3.1-8B is substantially less grounded (77--81\%), frequently generating abstract clinical labels (``lack of motivation,'' ``social withdrawal'') rather than verbatim transcript evidence.

Intra-model keyword stability follows the same rank ordering as ICC
(Gemma-9B~$>$~Phi-4~$>$~Llama-8B), confirming that score and reasoning
consistency are tightly coupled.  A consistent anxiety--depression
asymmetry emerges: keyword stability is universally lower for HADS-D
(intra Jaccard: 0.21--0.62) than for HADS-A (0.41--0.67), and
inter-model keyword agreement for depression (0.07--0.17) is roughly
half that for anxiety (0.22--0.25).  This aligns with depression
manifesting through latent lexical
features~\cite{stade2023,tausczik2010,alhanai2018} that are harder to
capture with consistent keywords than the more overt markers of
anxiety.

\begin{figure}[th!]
  \centering
  \includegraphics[width=\linewidth]{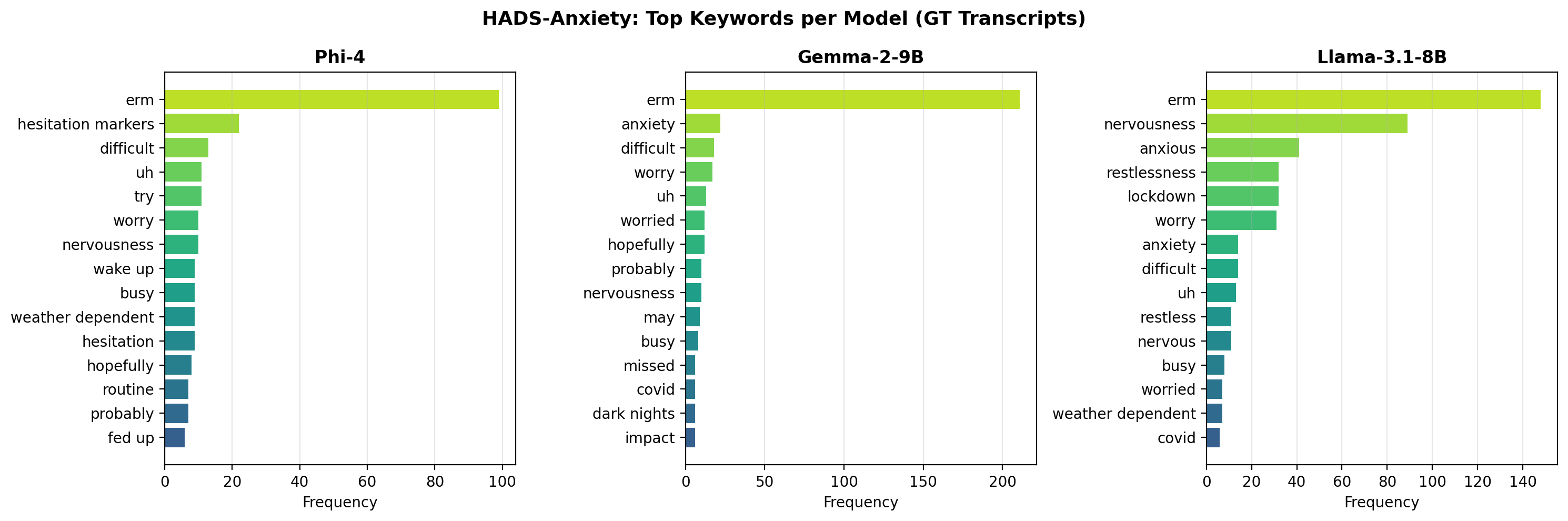}
  \caption{Top 15 keywords per model for HADS-Anxiety (GT transcripts). Note the dominance of ``erm'' across models and divergent keyword vocabularies underlying similar predictions.}
  \label{fig:keywords}
\end{figure}

The most striking finding is the score--evidence dissociation:
inter-model keyword Jaccard ranges from 0.07 to 0.25, yet score-level
$\scc$ reaches 0.87, a gap of 0.5--0.7 showing that models converge on
\textit{what} to predict far more than on \textit{why} they predict it
(Figure~\ref{fig:keywords}).%
\mnSL{The graphs in Figure \ref{fig:keywords} are too small. I wonder
  if a table wouldn't do the same job while being more readable.}
This implies that presenting a single model's keywords as
\textit{``the reason for this score''} overstates reasoning certainty;
only keywords corroborated across models or runs should be treated as
robust evidence~\cite{turpin2024,loweimi2025jstsp}.
\mnSL{Point well taken. Perhaps we could use similar measures to
  assess the value of LLM results in thematic analysis for the paper
  Sofia is leading.}

\begin{table}[t]
\caption{Keyword analysis (GT transcripts): groundedness, intra-model Jaccard (3 runs), and mean inter-model Jaccard (A\,=\,Anxiety, D\,=\,Depression).}
\label{tab:keywords}
\centering
\setlength{\tabcolsep}{3pt}
\begin{tabular}{@{}lcccccc@{}}
\toprule
& \multicolumn{2}{c}{Grounded.\ (\%)} & \multicolumn{2}{c}{Intra Jaccard} & \multicolumn{2}{c}{Inter Jaccard} \\
\cmidrule(lr){2-3}\cmidrule(lr){4-5}\cmidrule(lr){6-7}
Model & A & D & A & D & A & D \\
\midrule
Phi-4     & \textbf{93.0} & \textbf{94.3} & 0.48 & 0.38 & 0.25 & 0.15 \\
Gemma-9B  & \textbf{94.3} & \textbf{94.6} & \textbf{0.67} & \textbf{0.62} & 0.23 & 0.10 \\
Llama-8B  & 77.4 & 80.8 & 0.41 & 0.21 & 0.23 & 0.12 \\
\bottomrule
\end{tabular}
\end{table}

\section{Conclusions and Future Work}
\label{sec:conclusions}

We presented the first joint analysis of intra-model consistency, ASR robustness, and keyword evidence faithfulness for LLM-based mental health screening. Phi-4 and Gemma-2-9B demonstrate excellent consistency (ICC$>$0.89) and stable predictive validity ($\scc = 0.38$--$0.56$) across ASR conditions, whereas Llama-3.1-8B exhibits severe consistency degradation under ASR. Keyword analysis shows that reliable score predictions are accompanied by more grounded ($>$93\%) and stable evidence, but inter-model keyword agreement (Jaccard$\leq$0.25) remains far lower than score-level agreement ($\scc$ up to 0.87), revealing a pronounced score--evidence dissociation. These results establish model-specific ASR tolerance and evidence faithfulness as essential criteria for deployment and support fully automated Speech~$\to$~ASR~$\to$~LLM screening pipelines for models that satisfy all three reliability dimensions.

From a clinical perspective, the results indicate that reliability is best improved through model selection rather than aggressive ASR optimisation, and that three-run averaging is needed to mitigate stochastic variability. Evidence presented to clinicians should be automatically checked for groundedness and corroborated across runs or models, with particular caution for depression-related keywords.
We note that PsyVoiD was collected during a pandemic lockdown with a sex and age imbalance (63\% female, median age 62) and relatively few participants in the clinical severity range, which may limit generalisability.
Future work should evaluate broader ASR error regimes, additional ASR architectures, and more diverse populations, investigate whether stability converges beyond three runs or requires more, and examine whether multimodal approaches combining acoustic and textual features can improve robustness and interpretability for anxiety detection.


\bibliographystyle{IEEEtran}
\bibliography{ref}

@article{salari2020,
  author  = {Salari, N. and Hosseinian-Far, A. and Jalali, R. and Vaisi-Raygani, A. and Rasoulpoor, S. and Mohammadi, M. and Rasoulpoor, S. and Khaledi-Paveh, B.},
  title   = {Prevalence of stress, anxiety, depression among the general population during the {COVID-19} pandemic: a systematic review and meta-analysis},
  journal = {Globalization and Health},
  volume  = {16},
  pages   = {57},
  year    = {2020}
}

@article{mcgorry2018,
  author  = {McGorry, P. D. and Ratheesh, A. and O'Donoghue, B.},
  title   = {Early intervention---an implementation challenge for 21st century mental health care},
  journal = {JAMA Psychiatry},
  volume  = {75},
  number  = {6},
  pages   = {545--546},
  year    = {2018}
}

@article{vonkorff1987,
  author  = {Von Korff, M. and Shapiro, S. and Burke, J. D. and Teitlebaum, M. and Skinner, E. A. and German, P. and Turner, R. W. and Klein, L. and Burns, B.},
  title   = {Anxiety and depression in a primary care clinic},
  journal = {Archives of General Psychiatry},
  volume  = {44},
  number  = {2},
  pages   = {152--156},
  year    = {1987}
}

@article{zigmond1983,
  author  = {Zigmond, A. S. and Snaith, R. P.},
  title   = {The {Hospital Anxiety and Depression Scale}},
  journal = {Acta Psychiatrica Scandinavica},
  volume  = {67},
  number  = {6},
  pages   = {361--370},
  year    = {1983}
}

@article{newson2020,
  author  = {Newson, J. J. and Hunter, D. and Thiagarajan, T. C.},
  title   = {The heterogeneity of mental health assessment},
  journal = {Frontiers in Psychiatry},
  volume  = {11},
  pages   = {76},
  year    = {2020}
}

@article{cummins2015,
  author  = {Cummins, N. and Scherer, S. and Krajewski, J. and Schnieder, S. and Epps, J. and Quatieri, T. F.},
  title   = {A review of depression and suicide risk assessment using speech analysis},
  journal = {Speech Communication},
  volume  = {71},
  pages   = {10--49},
  year    = {2015}
}

@article{wu2011,
  author  = {Wu, S. and Falk, T. H. and Chan, W.-Y.},
  title   = {Automatic speech emotion recognition using modulation spectral features},
  journal = {Speech Communication},
  volume  = {53},
  number  = {5},
  pages   = {768--785},
  year    = {2011}
}

@article{low2020,
  author  = {Low, D. M. and Bentley, K. H. and Ghosh, S. S.},
  title   = {Automated assessment of psychiatric disorders using speech: a systematic review},
  journal = {Laryngoscope Investigative Otolaryngology},
  volume  = {5},
  number  = {1},
  pages   = {96--116},
  year    = {2020}
}

@inproceedings{loweimi2025,
  author    = {Loweimi, E. and de la Fuente Garcia, S. and Luz, S.},
  title     = {Zero-shot speech-based depression and anxiety assessment with {LLMs}},
  booktitle = {Proc. Interspeech},
  pages     = {489--493},
  year      = {2025}
}

@article{loweimi2025jstsp,
  author  = {Loweimi, E. and de la Fuente Garcia, S. and Luz, S.},
  title   = {Zero-shot speech-based mental health and affective state assessment using {LLMs}},
  journal = {IEEE J. Selected Topics in Signal Processing},
  note    = {Under review},
  year    = {2025}
}

@article{lan2024,
  author  = {Lan, X. and others},
  title   = {Depression detection on social media with large language models},
  journal = {arXiv preprint arXiv:2403.10750},
  year    = {2024}
}

@inproceedings{tao2023,
  author    = {Tao, Y. and others},
  title     = {Classifying anxiety and depression through {LLMs} virtual interactions: a case study with {ChatGPT}},
  booktitle = {Proc. IEEE BIBM},
  pages     = {2259--2264},
  year      = {2023}
}

@article{xu2025,
  author  = {Xu, S. and others},
  title   = {Identifying psychiatric manifestations in outpatients with depression and anxiety: a large language model-based approach},
  journal = {medRxiv},
  year    = {2025}
}

@article{lho2025,
  author  = {Lho, S. K. and others},
  title   = {Large language models and text embeddings for detecting depression and suicide in patient narratives},
  journal = {JAMA Network Open},
  volume  = {8},
  number  = {5},
  year    = {2025}
}

@article{guo2024,
  author  = {Guo, Z. and others},
  title   = {Large language models for mental health applications: systematic review},
  journal = {JMIR Mental Health},
  volume  = {11},
  number  = {1},
  pages   = {e57400},
  year    = {2024}
}

@article{liu2025,
  author  = {Liu, J. M. and others},
  title   = {Enhanced large language models for effective screening of depression and anxiety},
  journal = {arXiv preprint arXiv:2501.08769},
  year    = {2025}
}

@article{stade2023,
  author  = {Stade, E. C. and others},
  title   = {Depression and anxiety have distinct and overlapping language patterns: results from a clinical interview},
  journal = {J. Psychopathology and Clinical Science},
  year    = {2023}
}

@article{tausczik2010,
  author  = {Tausczik, Y. R. and Pennebaker, J. W.},
  title   = {The psychological meaning of words: {LIWC} and computerized text analysis methods},
  journal = {J. Language and Social Psychology},
  volume  = {29},
  number  = {1},
  pages   = {24--54},
  year    = {2010}
}

@article{guntuku2017,
  author  = {Guntuku, S. C. and Yaden, D. B. and Kern, M. L. and Ungar, L. H. and Eichstaedt, J. C.},
  title   = {Detecting depression and mental illness on social media: an integrative review},
  journal = {Current Opinion in Behavioral Sciences},
  volume  = {18},
  pages   = {43--49},
  year    = {2017}
}

@inproceedings{radford2023,
  author    = {Radford, A. and Kim, J. W. and Xu, T. and Brockman, G. and McLeavey, C. and Sutskever, I.},
  title     = {Robust speech recognition via large-scale weak supervision},
  booktitle = {Proc. ICML},
  year      = {2023}
}

@article{koo2016,
  author  = {Koo, T. K. and Li, M. Y.},
  title   = {A guideline of selecting and reporting intraclass correlation coefficients for reliability research},
  journal = {J. Chiropractic Medicine},
  volume  = {15},
  number  = {2},
  pages   = {155--163},
  year    = {2016}
}

@article{shrout1979,
  author  = {Shrout, P. E. and Fleiss, J. L.},
  title   = {Intraclass correlations: uses in assessing rater reliability},
  journal = {Psychological Bulletin},
  volume  = {86},
  number  = {2},
  pages   = {420--428},
  year    = {1979}
}

@misc{psyvoid2023,
  author = {de la Fuente Garcia, S. and Luz, S.},
  title  = {{PsyVoiD}---investigating the relationship between spontaneous speech features and psychology in the context of the {COVID-19} pandemic and lockdown},
  note   = {Dataset, University of Edinburgh},
  year   = {2023}
}

@inproceedings{brown2020,
  author    = {Brown, T. and others},
  title     = {Language models are few-shot learners},
  booktitle = {Advances in Neural Information Processing Systems},
  volume    = {33},
  pages     = {1877--1901},
  year      = {2020}
}

@inproceedings{alhanai2018,
  author    = {Al Hanai, T. and Ghassemi, M. and Glass, J.},
  title     = {Detecting depression with audio/text sequence modeling of interviews},
  booktitle = {Proc. Interspeech},
  pages     = {1716--1720},
  year      = {2018}
}

@inproceedings{delafuente2021,
  author    = {de la Fuente Garcia, S. and Haider, F. and Luz, S.},
  title     = {{COVID-19}: affect recognition through voice analysis during the winter lockdown in {Scotland}},
  booktitle = {Proc. IEEE EMBC},
  pages     = {2326--2329},
  year      = {2021}
}

@article{abdin2024,
  author  = {Abdin, M. and others},
  title   = {Phi-4 technical report},
  journal = {arXiv preprint arXiv:2412.08905},
  year    = {2024}
}

@article{gemma2024,
  author  = {{Google Team} and others},
  title   = {Gemma 2: improving open language models at a practical size},
  journal = {arXiv preprint arXiv:2408.00118},
  year    = {2024}
}

@article{grattafiori2024,
  author  = {Grattafiori, A. and others},
  title   = {The {Llama} 3 herd of models},
  journal = {arXiv preprint arXiv:2407.21783},
  year    = {2024}
}

@article{herrmann1997,
  author  = {Herrmann, C.},
  title   = {International experiences with the {Hospital Anxiety and Depression Scale}---a review of validation data and clinical results},
  journal = {J. Psychosomatic Research},
  volume  = {42},
  number  = {1},
  pages   = {17--41},
  year    = {1997}
}

@article{joyce2023,
  author  = {Joyce, D. W. and Aref-Adib, G. and Meyer, A. and Shivaprasad, B. and Abrahams, S.},
  title   = {Explainable artificial intelligence for mental health through transparency and interpretability for understandability},
  journal = {npj Digital Medicine},
  volume  = {6},
  number  = {6},
  year    = {2023}
}

@inproceedings{turpin2024,
  author    = {Turpin, M. and Michael, J. and Perez, E. and Bowman, S. R.},
  title     = {Language models don't always say what they think: unfaithful explanations in chain-of-thought prompting},
  booktitle = {Proc. NeurIPS},
  year      = {2024}
}

@article{wei2022,
  author  = {Wei, J. and others},
  title   = {Chain-of-thought prompting elicits reasoning in large language models},
  journal = {Advances in Neural Information Processing Systems},
  volume  = {35},
  pages   = {24824--24837},
  year    = {2022}
}

@article{eichstaedt2018,
  author  = {Eichstaedt, J. C. and others},
  title   = {Facebook language predicts depression in medical records},
  journal = {Proc. Natl Acad. Sci. USA},
  volume  = {115},
  number  = {44},
  pages   = {11203--11208},
  year    = {2018}
}

@article{bjelland2002,
  author  = {Bjelland, I. and Dahl, A. A. and Haug, T. T. and Neckelmann, D.},
  title   = {The validity of the {Hospital Anxiety and Depression Scale}: an updated literature review},
  journal = {J. Psychosomatic Research},
  volume  = {52},
  number  = {2},
  pages   = {69--77},
  year    = {2002}
}

@inproceedings{yang2024,
  author    = {Yang, K. and others},
  title     = {Towards interpretable mental health analysis with large language models},
  booktitle = {Proc. EMNLP},
  year      = {2024}
}

@inproceedings{ernala2019,
  author    = {Ernala, S. K. and others},
  title     = {Methodological gaps in predicting mental health states from social media},
  booktitle = {Proc. CHI},
  year      = {2019}
}

@article{malhotra2022,
  author  = {Malhotra, A. and Jindal, R.},
  title   = {Deep learning techniques for suicide and depression detection from online social media: a scoping review},
  journal = {Applied Soft Computing},
  volume  = {130},
  pages   = {109713},
  year    = {2022}
}

@inproceedings{lamichhane2023,
  author    = {Lamichhane, B.},
  title     = {Evaluation of {ChatGPT} for {NLP}-based mental health applications},
  booktitle = {arXiv preprint arXiv:2303.15727},
  year      = {2023}
}

@article{ji2023,
  author  = {Ji, Z. and others},
  title   = {Survey of hallucination in natural language generation},
  journal = {ACM Computing Surveys},
  volume  = {55},
  number  = {12},
  pages   = {1--38},
  year    = {2023}
}

@article{roy2025,
  author  = {Roy, K. and Surana, H. and Mullen, D. and Haut, K. and Flint, J. and Baxter, J.},
  title   = {Large language models for mental health diagnostic assessments: exploring the potential of {LLMs} for assisting with mental health diagnostic assessments},
  journal = {arXiv preprint arXiv:2501.01305},
  year    = {2025}
}

@article{rude2004,
  author  = {Rude, S. and Gortner, E.-M. and Pennebaker, J.},
  title   = {Language use of depressed and depression-vulnerable college students},
  journal = {Cognition \& Emotion},
  volume  = {18},
  number  = {8},
  pages   = {1121--1133},
  year    = {2004}
}

@article{calvo2017,
  author  = {Calvo, R. A. and Milne, D. N. and Hussain, M. S. and Christensen, H.},
  title   = {Natural language processing in mental health applications using non-clinical texts},
  journal = {Natural Language Engineering},
  volume  = {23},
  number  = {5},
  pages   = {649--685},
  year    = {2017}
}

\end{document}